\def\year{2017}\relax
\documentclass[letterpaper]{article}
\usepackage{aaai17_new}
\usepackage{times}
\usepackage{helvet}
\usepackage{courier}
\usepackage{graphicx}
\usepackage{algorithm} 
\usepackage{amsmath,amssymb,amsthm}
\usepackage{subfigure}
\usepackage{nicefrac}
\usepackage{url}

\usepackage{pifont}

\begin{document}
\title{A Deep Hierarchical Approach to Lifelong Learning in Minecraft}
\author{Chen Tessler$^*$, Shahar Givony$^*$, Tom Zahavy$^*$, Daniel J. Mankowitz$^*$, Shie Mannor \\
\small $^*$ equally contributed\\
Technion Israel Institute of Technology, Haifa, Israel\\
\small{chen.tessler, shahargiv, tomzahavy \{@campus.technion.ac.il \}, danielm@tx.technion.ac.il, shie@ee.technion.ac.il}}

\newtheorem{assume}{Assumption}
\newtheorem{define}{Definition}
\newtheorem{lemma}{Lemma}
\newtheorem{theorem}{Theorem}
\newtheorem{corollary}{Corollary}

\newcommand{\xmark}{\ding{55}}

\maketitle
\begin{abstract}
We propose a lifelong learning system that has the ability to reuse and transfer knowledge from one task to another while efficiently retaining the previously learned knowledge-base. Knowledge is transferred by learning reusable skills to solve tasks in Minecraft, a popular video game which is an unsolved and high-dimensional lifelong learning problem. These reusable skills, which we refer to as Deep Skill Networks, are then incorporated into our novel Hierarchical Deep Reinforcement Learning Network (H-DRLN) architecture using two techniques: (1) a deep skill array and (2) skill distillation, our novel variation of policy distillation \cite{Rusu2015} for learning skills. Skill distillation enables the H-DRLN to efficiently retain knowledge and therefore scale in lifelong learning, by accumulating knowledge and encapsulating multiple reusable skills into a single distilled network. The H-DRLN exhibits superior performance and lower learning sample complexity compared to the regular Deep Q Network \cite{Mnih2015} in sub-domains of Minecraft. 
\end{abstract}

\section{Introduction}
Lifelong learning considers systems that continually learn new tasks, from one or more domains, over the course of a lifetime. Lifelong learning is a large, open problem and is of great importance to the development of general purpose Artificially Intelligent (AI) agents.  A formal definition of lifelong learning follows.\\

\begin{define}
\label{def:lifelong}
Lifelong Learning is the continued learning of tasks, from one or more domains, over the course of a lifetime, by a lifelong learning system. A lifelong learning system efficiently and effectively (1) retains the knowledge it has learned; (2) selectively transfers knowledge to learn new tasks; and (3) ensures the effective and efficient interaction between (1) and (2)\cite{Silver2013}.\\
\end{define}

\begin{figure}
\begin{center}
\includegraphics[width=0.47\textwidth]{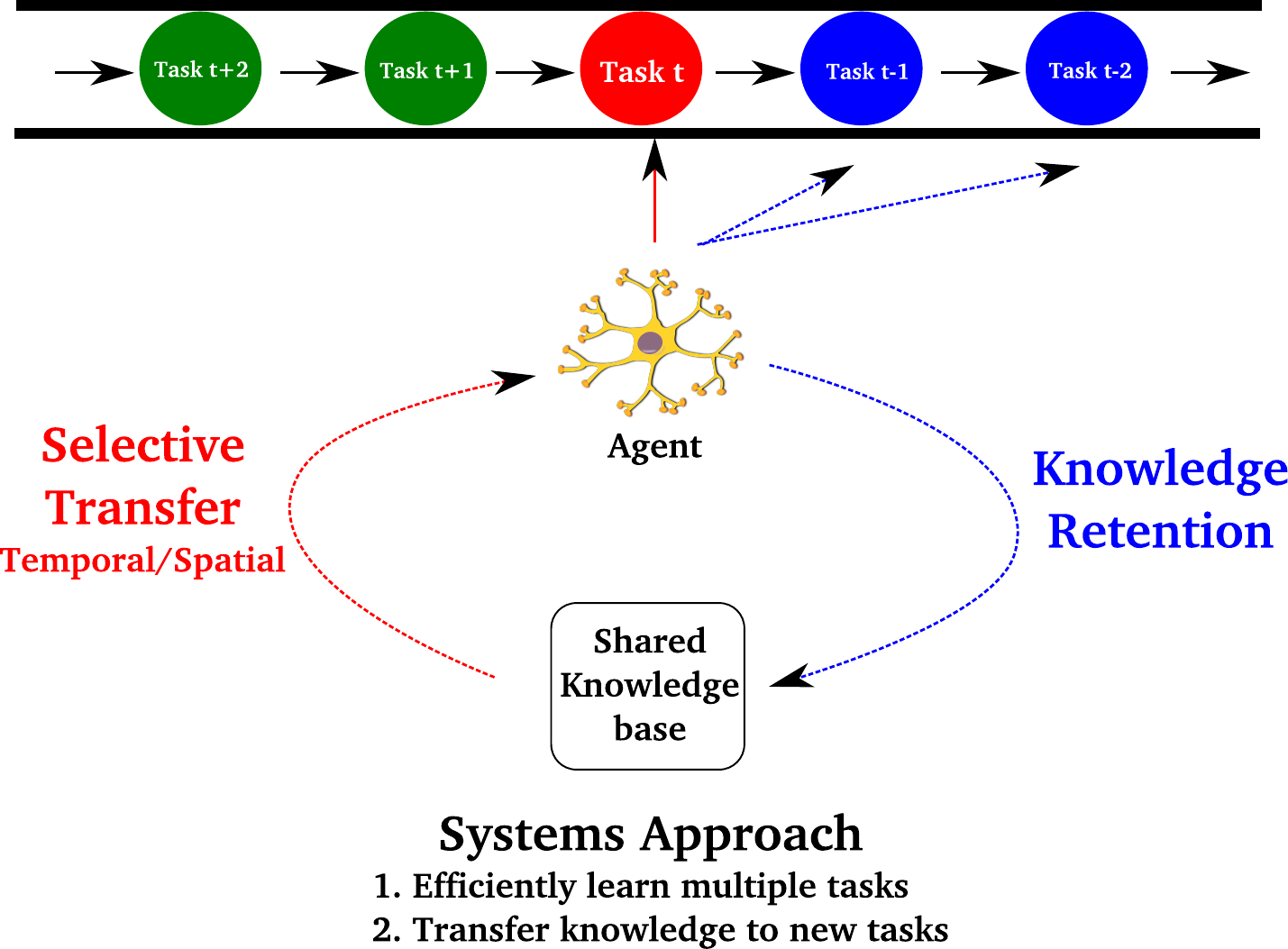} 
\caption{\textbf{Lifelong Learning:} A lifelong learning system (1) efficiently retains knowledge and (2) selectively transfers knowledge to solve new tasks. Upon solving a task, the knowledge base is refined and new knowledge is added to the system. A systems approach ensures efficient and effective interaction between (1) and (2).}
\label{fig:lls}
\end{center}
\end{figure}

A truly general lifelong learning system, shown in Figure \ref{fig:lls}, therefore has the following attributes:
(1) \textbf{Efficient retention of learned task knowledge}
A lifelong learning system should minimize the retention of erroneous knowledge. 
In addition, it should also be computationally efficient when storing knowledge in long-term memory. 
(2) \textbf{Selective transfer: } A lifelong learning system needs the ability to choose relevant prior knowledge for solving new tasks, while casting aside irrelevant or obsolete information.
(3) \textbf{System approach: } Ensures the effective and efficient interaction of the retention and transfer elements.\\

\begin{figure}
\begin{center}
\includegraphics[width=0.46\textwidth]{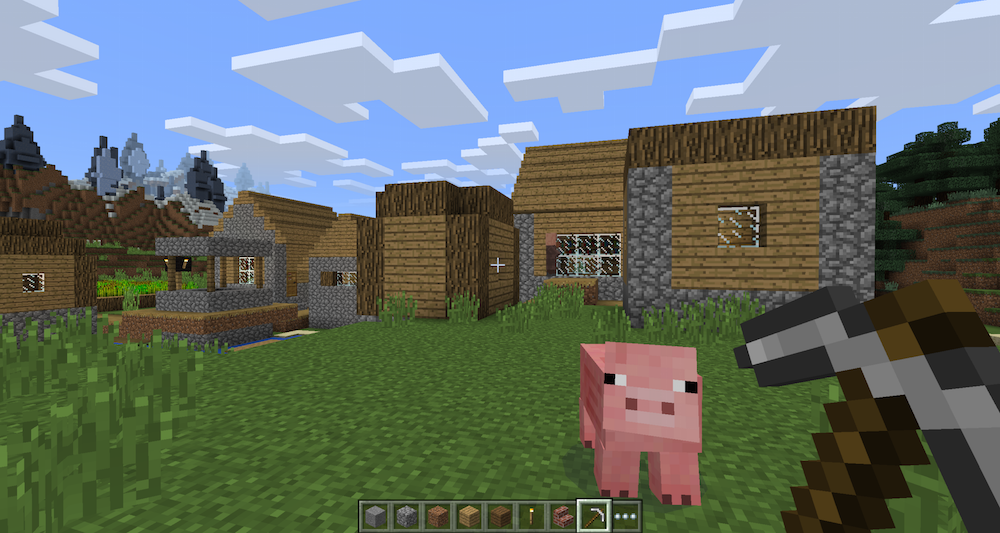} 
\caption{A screenshot from \textbf{Minecraft}, a popular video game which poses a challenging lifelong learning problem.}
\label{fig:minecraft}
\end{center}
\end{figure}

Lifelong learning systems in real-world domains suffer from the curse of dimensionality. That is, as the state and action spaces increase, it becomes more and more difficult to model and solve new tasks as they are encountered. In addition, planning over potentially infinite time-horizons as well as efficiently retaining and reusing knowledge pose non-trivial challenges.  A challenging, high-dimensional domain that incorporates many of the elements found in lifelong learning is Minecraft. Minecraft is a popular video game whose goal is to build structures, travel on adventures, hunt for food and avoid zombies. An example screenshot from the game is seen in Figure \ref{fig:minecraft}. Minecraft is an open research problem as it is impossible to solve the entire game using a single AI technique \cite{Smith2016,Oh2016}. Instead, the solution to Minecraft may lie in solving sub-problems, using a divide-and-conquer approach, and then providing a synergy between the various solutions. Once an agent learns to solve a sub-problem, it has acquired a \textit{skill} that can then be reused when a similar sub-problem is subsequently encountered. \\

\vspace{1cm}
Many of the tasks that are encountered by an agent in a lifelong learning setting can be naturally decomposed into \textit{skill hierarchies} \cite{Stone2000,Stone2005,Bai2015}. In Minecraft for example, consider building a wooden house as seen in Figure \ref{fig:minecraft}. This task can be decomposed into sub-tasks (a.k.a skills) such as chopping trees, sanding the wood, cutting the wood into boards and finally nailing the boards together. Here, the knowledge gained from chopping trees can also be partially \textit{reused} when cutting the wood into boards. In addition, if the agent receives a new task to build a small city, then the agent can reuse the \textit{skills} it acquired during the `building a house' task.\\

In a high-dimensional, lifelong learning setting such as Minecraft, learning skills and when to reuse the skills is non-trivial. This is key to efficient knowledge retention and transfer, increasing exploration, efficiently solving tasks and ultimately advancing the capabilities of the Minecraft agent. \\

Reinforcement Learning (RL) provides a generalized approach to skill learning through the options framework \cite{Sutton1999}. Options are Temporally Extended Actions (TEAs) and are also referred to as skills \cite{daSilva2012} and macro-actions \cite{Hauskrecht1998}. Options have been shown both theoretically \cite{Precup1997,Sutton1999} and experimentally \cite{Mann2013,Mann2014b} to speed up the convergence rate of RL algorithms. From here on in, we will refer to options as skills.\\

\vspace{1cm}
In order to learn reusable skills in a lifelong learning setting, the framework needs to be able to (1) learn skills, (2) learn a controller which determines when a skill should be used and \textit{reused} and (3) be 
able to efficiently accumulate reusable skills.
There are recent works that perform skill learning \cite{Mankowitz2016a,Mankowitz2016b,Mnih2016,Bacon2015}, but these works have focused on learning good skills and have not explicitly shown the ability to reuse skills nor scale with respect to the number of skills in lifelong learning domains.\\

With the emergence of Deep RL, specifically Deep Q-Networks (DQNs), RL agents are now equipped with a powerful non-linear function approximator that can learn rich and complex policies (or skills). Using these networks the agent learns policies (or skills) from raw image pixels, requiring less domain specific knowledge to solve complicated tasks (E.g Atari video games). While different variations of the DQN algorithm exist \cite{Van2015,schaul2015prioritized,wang2015dueling,bellemare2015increasing}, we will refer to the vanilla version unless otherwise stated. There are deep learning approaches that perform sub-goal learning \cite{Rusu2016,Kulkarni2016}, yet these approaches rely on providing  the task or sub-goal to the agent, prior to making a decision. \citeauthor{Kulkarni2016} (\citeyear{Kulkarni2016}) also rely on manually constructing sub-goals a-priori for tasks and utilize intrinsic motivation which may be problematic for complicated problems where designing good intrinsic motivations is not clear and non-trivial.  \\

In our paper, we present  our novel lifelong learning system called the Hierarchical Deep Reinforcement Learning (RL) Network (H-DRLN) architecture shown in Figure \ref{fig:hdrln} (It is defined formally in the Hierarchical Deep RL Network Section).  While we do not claim to provide an end-to-end solution, the H-DRLN contains all the basic building blocks of a truly general lifelong learning framework (see the Related Work Section for an in-depth overview). The H-DRLN controller learns to solve complicated tasks in Minecraft by learning reusable RL skills in the form of pre-trained Deep Skill Networks (DSNs). Knowledge is \textbf{retained} by incorporating reusable skills into the H-DRLN via a Deep Skill module. There are two types of Deep Skill Modules: (1) a DSN array (Figure \ref{fig:hdrln}, Module $A$) and (2) a \textit{multi-skill distillation} network (Figure \ref{fig:hdrln}, Module $B$), our novel variation of policy distillation \cite{Rusu2015} applied to learning skills. Multi-skill distillation enables the H-DRLN to \textit{efficiently} retain knowledge and therefore scale in lifelong learning, by encapsulating multiple reusable skills into a single distilled network. When solving a new task, the H-DRLN \textbf{selectively transfers} knowledge in the form of temporal abstractions (skills) to solve the given task. By taking advantage of temporally extended actions (skills), the H-DRLN learns to solve tasks with lower sample complexity and superior performance compared to vanilla DQNs. \\

\textbf{Main Contributions:} (1) A novel Hierarchical Deep Reinforcement Learning Network (H-DRLN) architecture which includes an H-DRLN controller and a Deep Skill Module. The H-DRLN contains all the basic building blocks for a truly general lifelong learning framework. (2) We show the potential to learn \textit{reusable} Deep Skill Networks (DSNs) and perform knowledge transfer of the learned DSNs to new tasks to obtain an optimal solution. We also show the  potential to transfer knowledge between related tasks without any additional learning. (3) We efficiently retain knowledge in the H-DRLN by performing skill distillation, our variation of policy distillation, for learning skills and incorporate it into the Deep Skill Model to solve complicated tasks in Minecraft. (4) Empirical results for learning an H-DRLN in sub-domains of Minecraft with a DSN array and a distilled skill network. We also verify the improved convergence guarantees for utilizing reusable DSNs (a.k.a options) within the H-DRLN, compared to the vanilla DQN. 


\section{Previous Research on Lifelong Learning in RL}
\label{sec:related_work}

Designing a truly general lifelong learning agent is a challenging task. Previous works on lifelong learning in RL have focused on solving specific elements of the general lifelong learning system as shown in Table \ref{tab:lifelong}. \\

According to Definition \ref{def:lifelong}, a lifelong learning agent should be able to \textbf{efficiently retain knowledge}. This is typically done by sharing a representation among tasks, using distillation \cite{Rusu2015} or a latent basis \cite{Ammar2014}. The agent should also learn to \textbf{selectively use} its past knowledge to solve new tasks efficiently. Most works have focused on a \textit{spatial transfer} mechanism, i.e., they suggested learning differentiable weights from a shared representation to the new tasks \cite{2016arXiv161105397J,Rusu2016}. In contrast, \citeauthor{Brunskill2014} (\citeyear{Brunskill2014}) suggested a \textit{temporal transfer} mechanism, which identifies an optimal set of skills in past tasks and then learns to use these skills in new tasks. Finally, the agent should have a \textbf{systems approach} that allows it to efficiently retain the knowledge of \textit{multiple tasks} as well as an efficient mechanism to \textit{transfer} knowledge for solving new tasks.\\

Our work incorporates all of the basic building blocks necessary to performing lifelong learning. As per the lifelong learning definition, we efficiently transfer knowledge from previous tasks to solve a new target task by utilizing RL skills \cite{Sutton1999}. We show that skills reduce the sample complexity in a complex Minecraft environment and suggest an efficient mechanism to retain the knowledge of multiple skills that is scalable with the number of skills. 

\begin{table}
\scalebox{0.6}{
\begin{tabular}{|c|c|c|c|c|c|c|c|}
\hline 
 & \textbf{Works} & H-DRLN & Ammar & Brunskill  & Rusu & Rusu  & Jaderberg \tabularnewline
 &  & (this work) &  et. al & and Li &  & et. al. & et. al.\tabularnewline
\textbf{Attribute} &  &  & (2014) & (2014) & (2015) & (2016) & (2016)\tabularnewline
\hline 
 & Memory &  &  &  &  &  & \tabularnewline
 &  efficient & \checkmark & \checkmark & \xmark & \checkmark & \xmark & \xmark\tabularnewline
\textbf{Knowledge} & architecture &  &  &  &  &  & \tabularnewline
\cline{2-8} 
\textbf{Retention} & Scalable to &  &  &  &  &  & \tabularnewline
 & high & \checkmark & \xmark & \xmark & \checkmark & \checkmark & \checkmark\tabularnewline
 & dimensions &  &  &  &  &  & \tabularnewline
\hline 
 & Temporal  &  &  &  &  &  & \tabularnewline
 & abstraction & \checkmark & \xmark & \checkmark & \xmark & \xmark & \xmark\tabularnewline
\textbf{Selective} & transfer &  &  &  &  &  & \tabularnewline
\cline{2-8} 
\textbf{Transfer} & Spatial &  &  &  &  &  & \tabularnewline
 & abstraction & \xmark & \checkmark & \xmark & \checkmark & \checkmark & \checkmark\tabularnewline
 & transfer &  &  &  &  &  & \tabularnewline
\hline 
 &  &  &  &  &  &  & \tabularnewline
 & Multi-task & \checkmark & \checkmark & \checkmark & \checkmark & \xmark & \checkmark\tabularnewline
\textbf{Systems} &  &  &  &  &  &  & \tabularnewline
\cline{2-8} 
\textbf{Approach} &  &  &  &  &  &  & \tabularnewline
 & Transfer & \checkmark & \checkmark & \checkmark & \xmark & \checkmark & \checkmark\tabularnewline
 &  &  &  &  &  &  & \tabularnewline
\hline 
\end{tabular}
}
\caption{\textbf{Previous works} on lifelong learning in RL.}
\label{tab:lifelong}
\end{table}

\section{Background}
\label{sec:background}

\textbf{Reinforcement Learning:} The goal of an RL agent is to maximize its expected return by learning a policy $\pi:S \rightarrow \Delta_A$ which is a mapping from states $s \in S$ to a probability distribution over the actions $A$. At time $t$ the agent observes a state $s_t \in S$, selects an action $a_t \in A$, and receives a bounded reward $r_t \in [0, R_{\max}]$ where $R_{\max}$ is the maximum attainable reward and $\gamma\in[0,1]$ is the discount factor. Following the agents action choice, it transitions to the next state $s_{t+1} \in S$ . We consider infinite horizon problems where the cumulative return at time $t$ is given by $R_t = \sum_{t'=t}^\infty \gamma^{t'-t}r_t$. The action-value function $Q^{\pi}(s,a) = \mathbb{E} [R_t|s_t = s, a_t = a, \pi]$ represents the expected return after observing state $s$ and taking an action $a$ under
a policy $\pi$. The optimal action-value function obeys a fundamental recursion known as the Bellman equation:

\begin{equation*}
Q^* (s_t,a_t)=\mathbb{E}
\left[r_t+\gamma \underset{a'}{\mathrm{max}}Q^*(s_{t+1},a')
\right] \enspace .
\end{equation*}

\textbf{Deep Q Networks:} The DQN algorithm \cite{Mnih2015} approximates the optimal Q function with a Convolutional Neural Network (CNN) \cite{Krizhevsky2012}, by optimizing the network weights such that the expected Temporal Difference (TD) error of the optimal bellman equation (Equation \ref{DQN_loss}) is minimized:

\begin{equation}
\label{DQN_loss}
\mathbb{E}_{s_t,a_t,r_t,s_{t+1}}\left\Vert Q_{\theta}\left(s_{t},a_{t}\right)-y_{t}\right\Vert _{2}^{2} \enspace ,
\end{equation}

where 

\begin{equation*}
y_{t}=
\begin{cases}
r_{t} & \mbox{if } s_{t+1} \mbox{ is terminal}\\
r_{t}+\gamma\underset{\mbox{\mbox{\ensuremath{a}'}}}{\mbox{max}}Q_{\theta_{target}}\left(s_{t+1},a^{'}\right) & \mbox{otherwise}
\end{cases}
\end{equation*}
 
Notice that this is an offline learning algorithm, meaning that the tuples $\left\{ s_{t,}a_{t},r_{t},s_{t+1},\gamma\right\}$ are collected from the agents experience and are stored in the \textbf{Experience Replay (ER)} \cite{lin1993reinforcement}. The ER is a buffer that stores the agent’s experiences at each time-step $t$, for the purpose of ultimately training the DQN parameters to minimize the loss function. When we apply minibatch training updates, we sample tuples of experience at random from the pool of stored samples in the ER. The DQN maintains two separate Q-networks. The current Q-network with parameters $\theta$, and the target Q-network with parameters $\theta_{target}$. The parameters $\theta_{target}$ are set to $\theta$ every fixed number of iterations. In order to capture the game dynamics, the DQN represents the state by a sequence of image frames.\\

\textbf{Double DQN \cite{Van2015}: } Double DQN (DDQN) prevents overly optimistic estimates of the value function. This is achieved by performing action selection with the current network $\theta$ and evaluating the action with the target network $\theta_{target}$ yielding the DDQN target update $y_t=r_t$ if $s_{t+1}$ is terminal, otherwise $y_t = r_t + \gamma Q_{\theta_{target}}(s_{t+1}, \max_{a} Q_{\theta}(s_{t+1}, a))$. DDQN is utilized in this paper to improve learning performance.\\

\textbf{Skills, Options, Macro-actions \cite{Sutton1999}:} A skill $\sigma$ is a temporally extended control structure defined by a triple $\sigma = <I,\pi,\beta>$ where I is the set of states where the skill can be initiated, $\pi$ is the intra-skill policy, which determines how the skill behaves in encountered states, and $\beta$ is the set of termination probabilities determining when a skill will stop executing. The parameter $\beta$ is typically a function of state $s$ or time $t$.\\

\textbf{Semi-Markov Decision Process (SMDP):} Planning with skills can be performed using SMDP theory.  More formally, an SMDP can be defined by a five-tuple $<S, \Sigma, P, R, \gamma>$ where $S$ is a set of states, $\Sigma$ is a set of skills, and $P$ is the transition probability kernel. We assume rewards received at each timestep are bounded by $[0, R_{\max}]$. $R:S \times \sigma \rightarrow [0,\frac{R_{\max}}{1-\gamma}]$ represents the expected discounted sum of rewards received during the execution of a skill $\sigma$ initialized from a state $s$. The solution to an SMDP is a skill policy $\mu$.\\

\textbf{Skill Policy:} A skill policy $\mu : S \rightarrow \Delta_\Sigma$  is a mapping from states
to a probability distribution over skills $\Sigma$. The action-value function $Q_µ : S \times \Sigma \rightarrow R$ represents the long-term value of taking a skill $\sigma \in \Sigma$ from a state $s \in S$ and thereafter always selecting skills according to policy $\mu$ and is defined by $Q_µ(s, \sigma) = \mathbb{E} [\sum ^\infty _{t=0} \gamma ^t R_t |(s, \sigma), \mu] $.
We denote the skill reward as $R_s^{\sigma} = \mathbb{E}[r_{t+1} + \gamma r_{t+2} + \cdot\cdot\cdot + \gamma ^{k-1} r_{t+k} | s_t=s,\sigma]$  and transition probability as  $P_{s,s'}^{\sigma} = \sum_{j=0}^\infty \gamma^j Pr[k=j,s_{t+j}=s'|s_t=s,\sigma]$. Under these definitions the optimal skill value function is given by the following equation \cite{stolle2002learning}:
\begin{equation}
\label{OptionBellman}
Q_{\Sigma}^*(s,\sigma) = \mathbb{E} [R_s^{\sigma} + \gamma ^k \underset{\sigma'\in \Sigma}{\mathrm{max}} Q_{\Sigma}^*(s',\sigma')] \enspace .
\end{equation}

\textbf{Policy Distillation \cite{Rusu2015}: }
Distillation \cite{hinton2015distilling} is a method to transfer knowledge from a teacher model $T$ to a student model S. This process is typically done by supervised learning. For example, when both the teacher and the student are separate deep neural networks, the student network is trained to predict the teacher's output layer (which acts as labels for the student). Different objective functions have been previously proposed. In this paper we input the teacher output into a softmax function and train the distilled network using the Mean-Squared-Error (MSE) loss: $\mbox{cost}(s)=\Vert \mbox{Softmax}_\tau (Q_{T}(s)) - Q_{S}(s)\Vert^2$ where $Q_{T}(s)$ and $Q_{S}(s)$ are the action values of the teacher and student networks respectively and $\tau$ is the softmax temperature. During training, this cost function is differentiated according to the student network weights.

Policy distillation can be used to transfer knowledge from $N$ teachers $T_i, i=1,\cdots N$ into a single student (multi-task policy distillation). This is typically done by switching between the $N$ teachers every fixed number of iterations during the training process. When the student is learning from multiple teachers (i.e., multiple policies), a separate student output layer is assigned to each teacher $T_i$, and is trained for each task, while the other layers are shared.

\section{Hierarchical Deep RL Network}

\begin{figure}[t]
\begin{center}
\includegraphics[width=0.45\textwidth]{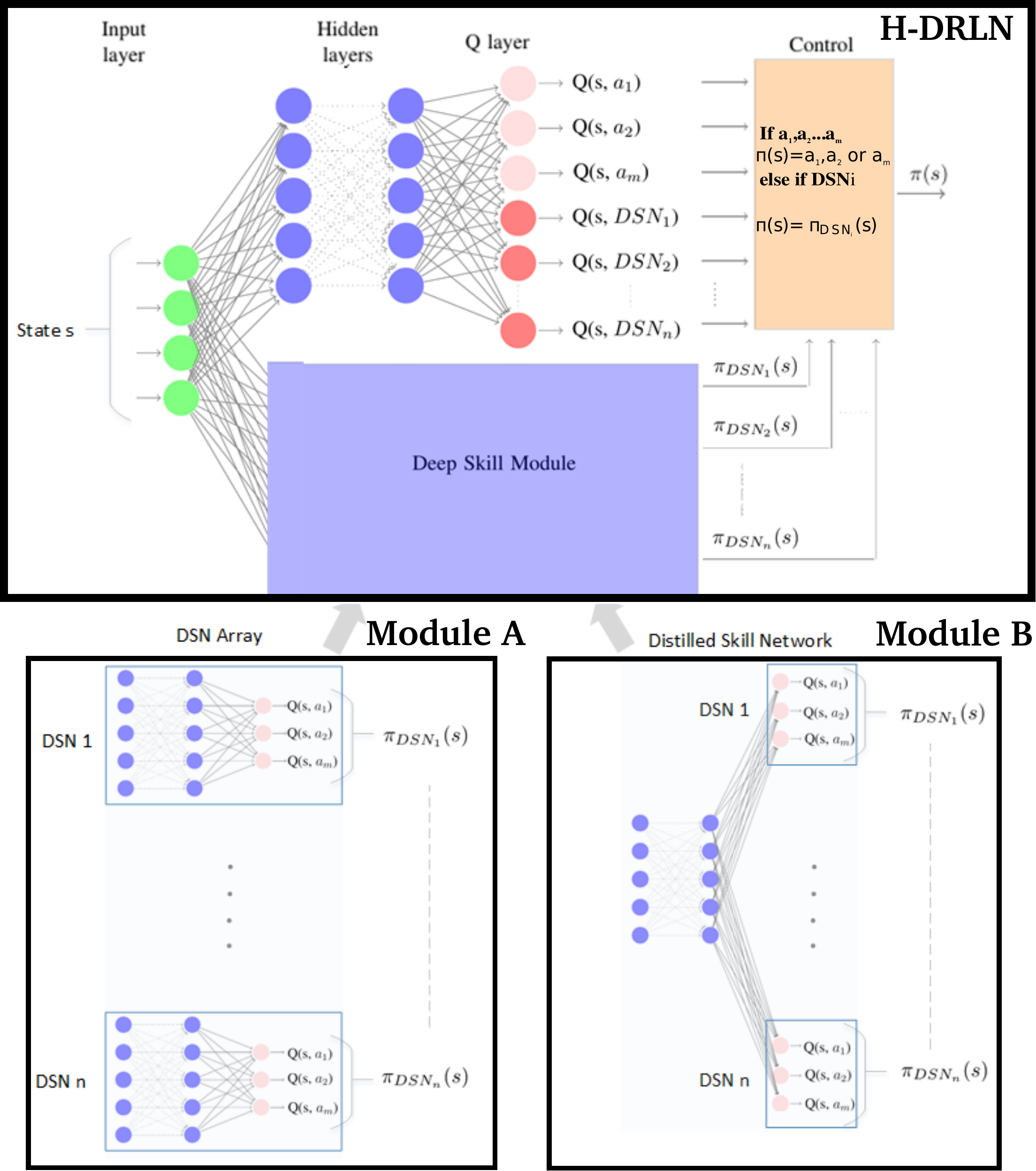} 
\caption{\textbf{The H-DRLN architecture:} It has outputs that correspond to primitive actions ($a_1,a_2,...,a_m$) and DSNs ($DSN_1, DSN_2, ... ,DSN_n$). The Deep Skill Module (bottom) represents a set of skills. It receives an input and a skill index and outputs an action according to the corresponding skill policy. The architecture of the deep skill module can be either a DSN array or a Distilled Multi-Skill Network. }
\label{fig:hdrln}
\end{center}
\end{figure}

In this Section, we present an in-depth description of the H-DRLN (Figure \ref{fig:hdrln}); a new architecture that extends the DQN and facilitates skill reuse in lifelong learning. Skills are incorporated into the H-DRLN via a Deep Skill Module that can incorporate either a DSN array or a distilled multi-skill network.\\

\textbf{Deep Skill Module: }
The pre-learned skills are represented as deep networks and are referred to as Deep Skill Networks (DSNs). They are trained a-priori on various sub-tasks using our version of the DQN algorithm and the regular Experience Replay (ER) as detailed in the Background Section. Note that the DQN is one choice of architecture and, in principal, other suitable networks may be used in its place. The Deep Skill Module represents a set of $N$ DSNs. Given an input state $s \in S$ and a skill index $i$, it outputs an action $a$ according to the corresponding DSN policy $\pi_{DSN_{i}}$. We propose two different Deep Skill Module architectures: (1) The DSN Array (Figure \ref{fig:hdrln},  module $A$): an array of pre-trained DSNs where each DSN is represented by a separate DQN. (2) The Distilled Multi-Skill Network (Figure \ref{fig:hdrln}, module $B$), a single deep network that represents multiple DSNs. Here, the different DSNs share all of the hidden layers while a separate output layer is trained for each DSN via policy distillation \cite{Rusu2015}. The Distilled skill network allows us to incorporate multiple skills into a single network, making our architecture scalable to lifelong learning with respect to the number of skills.\\

\textbf{H-DRLN architecture: } A diagram of the H-DRLN architecture is presented in Figure \ref{fig:hdrln} (top). Here, the outputs of the H-DRLN consist of primitive actions as well as skills. The H-DRLN learns a policy that determines when to execute primitive actions and when to \textbf{reuse} pre-learned skills. If the H-DRLN chooses to execute a primitive action $a_t$ at time $t$, then the action is executed for a single timestep. However, if the H-DRLN chooses to execute a skill $\sigma_i$ (and therefore DSN $i$ as shown in Figure \ref{fig:hdrln}), then DSN $i$ executes its policy, $\pi_{DSN_{i}}(s)$ until it terminates and then gives control back to the H-DRLN. This gives rise to two necessary modifications that we needed to make in order to incorporate skills into the learning procedure and generate a truly hierarchical deep network: (1) Optimize an objective function that incorporates skills; (2) Construct an ER that stores skill experiences. \\

\textbf{Skill Objective Function:} As mentioned previously, a H-DRLN extends the vanilla DQN architecture to learn control between primitive actions and skills. The H-DRLN loss function has the same structure as Equation~\ref{DQN_loss}, however instead of minimizing the standard Bellman equation, we minimize the Skill Bellman equation (Equation~\ref{OptionBellman}). More specifically, for a skill $\sigma_t$ initiated in state $s_t$ at time $t$ that has executed for a duration $k$, the H-DRLN target function is given by: 

\begin{equation}\nonumber
\resizebox{0.45\textwidth}{!}{$
y_{t}=
\begin{cases}
\sum_{j=0}^{k-1} \left[ \gamma^j r_{j+t} \right] & \mbox{if } s_{t+k} \mbox{ terminal}\\
 \sum_{j=0}^{k-1} \left[ \gamma^j r_{j+t} \right]+

 \gamma^k\underset{\mbox{\mbox{\ensuremath{\sigma}'}}}{\mbox{max}}Q_{\theta_{target}}\left(s_{t+k},\sigma^{'}\right) & \mbox{else}
\end{cases} 
$}
\end{equation}

This is the first work to incorporate an SMDP cost function into a deep RL setting.\\

\textbf{Skill - Experience Replay: } We extend the regular ER \cite{lin1993reinforcement} to incorporate skills and term this the Skill Experience Replay (S-ER). 
There are two differences between the standard ER and our S-ER. Firstly, for each sampled skill tuple, we calculate the sum of discounted cumulative rewards, $\tilde{r}$, generated whilst executing the skill. Second, since the skill is executed for $k$ timesteps, we store the transition to state $s_{t+k}$ rather than $s_{t+1}$. This yields the skill tuple $(s_t,\sigma_t, \tilde{r}_t, s_{t+k})$ where $\sigma_t$ is the skill executed at time $t$.

\section{Experiments}
\label{sec:experiments}

\begin{figure}[h]
\begin{center}

\includegraphics[width=0.40\textwidth]{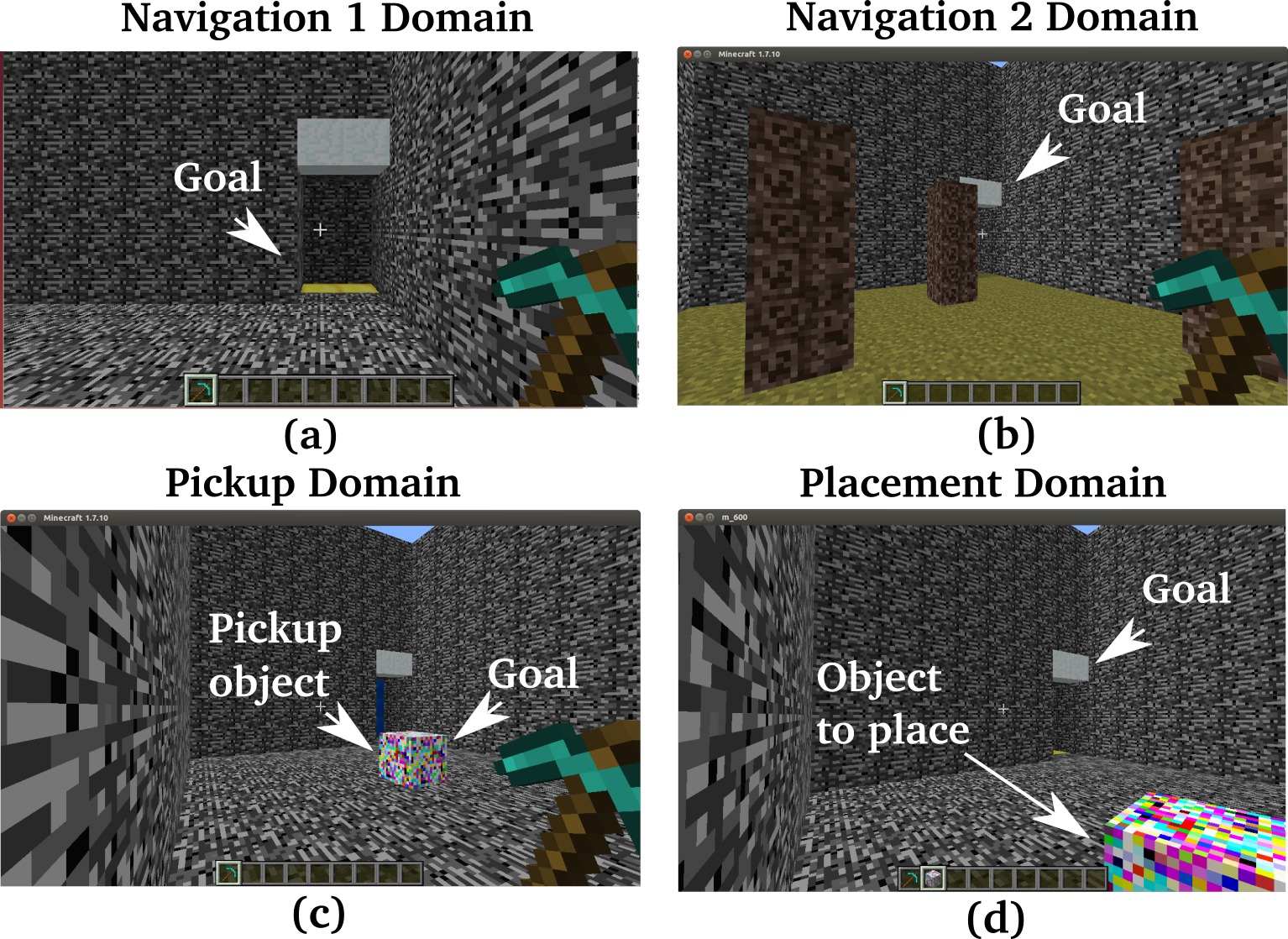} 
\caption{\textbf{The domains:} ($a$)-($d$) are screenshots for each of the domains we used to train the DSNs.}
\label{fig:domains_dsn}
\end{center}
\end{figure}

To solve new tasks as they are encountered in a lifelong learning scenario, the agent needs to be able to adapt to new game dynamics and learn when to \textit{reuse} skills that it has learned from solving previous tasks. In our experiments, we show (1) the ability of the Minecraft agent to learn DSNs on sub-domains of Minecraft (shown in Figure \ref{fig:domains_dsn}$a-d$). (2) The ability of the agent to reuse a DSN from navigation domain 1 (Figure \ref{fig:domains_dsn}$a$) to solve a new and more complex task, termed the \textit{two-room} domain (Figure \ref{fig:domains_complex}$a$). (3) The potential to transfer knowledge between related tasks without any additional learning. (4) We demonstrate the ability of the agent to reuse multiple DSNs to solve the \textit{complex-domain} (Figure \ref{fig:domains_complex}$b$). (5) We use two different Deep Skill Modules and demonstrate that our architecture scales for lifelong learning.\\

\textbf{State space} - As in \citeauthor{Mnih2015} (\citeyear{Mnih2015}), the state space is represented as raw image pixels from the last four image frames which are combined and down-sampled into an $84 \times 84$ pixel image.  \textbf{Actions} - The primitive action space for the DSN consists of six actions: (1) Move forward, (2) Rotate left by $30^{\circ}$, (3) Rotate right by $30^{\circ}$, (4) Break a block, (5) Pick up an item and (6) Place it. \textbf{Rewards} - In all domains, the agent gets a small negative reward signal after each step and a non-negative reward upon reaching the \textbf{final} goal (See Figure \ref{fig:domains_dsn} and Figure \ref{fig:domains_complex} for the different domain goals). \\

\textbf{Training} - Episode lengths are $30, 60$ and $100$ steps for single DSNs, the two room domain and the complex domain respectively. The agent is initialized in a random location in each DSN and in the first room for the two room and complex domains. \textbf{Evaluation} - the agent is evaluated during training using the current learned architecture every 20k (5k) optimization steps (a single epoch) for the DSNs (two room and complex room domains). During evaluation, we averaged the agent's performance over 500 (1000) steps respectively. \textbf{Success percentage: } The $\%$ of successful task completions during evaluation.\\

\subsection{Training a DSN}

Our first experiment involved training DSNs in sub-domains of Minecraft (Figure \ref{fig:domains_dsn}$a-d$), including two navigation domains, a pickup domain and a placement domain respectively. The break domain is the same as the placement domain, except it ends with the break action. Each of these domains come with different learning challenges. The Navigation $1$ domain is built with identical walls, which provides a significant learning challenge since there are visual ambiguities with respect to the agent's location (see Figure \ref{fig:domains_dsn}$a$). The Navigation $2$ domain provides a different learning challenge since there are obstacles that occlude the agent's view of the exit from different regions in the room (Figure \ref{fig:domains_dsn}$b$). The pick up (Figure \ref{fig:domains_dsn}$c$), break and placement (Figure \ref{fig:domains_dsn}$d$) domains require navigating to a specific location and ending with the execution of a primitive action (Pickup, Break or Place respectively).  \\

In order to train the different DSNs, we use the Vanilla DQN architecture \cite{Mnih2015} and performed a grid search to find the optimal hyper-parameters for learning DSNs in Minecraft. The best parameter settings that we found include: (1) a higher learning ratio (iterations between emulator states, \textit{n-replay} = 16), (2) higher learning rate (\textit{learning rate} = 0.0025) and (3) less exploration (\textit{eps{\_}endt} - 400K). We implemented these modifications, since the standard Minecraft emulator has a slow frame rate (approximately $400$ ms per emulator timestep), and these modifications enabled the agent to increase its learning between game states. We also found that a smaller experience replay (\textit{replay{\_}memory} - 100K) provided improved performance, probably due to our task having a relatively short time horizon (approximately $30$ timesteps). The rest of the parameters from the Vanilla DQN remained unchanged. After we tuned the hyper-parameters, all the DSNs managed to solve the corresponding sub-domains with almost $100\%$ success as shown in Table \ref{table:distilled}. (see supplementary material for learning curves).

\subsection{Training an H-DRLN with a DSN}
In this experiment, we train the H-DRLN agent to solve a complex task, the two-room domain, by reusing a single DSN (pre-trained on the navigation $1$ domain). \\

\begin{figure}
\begin{center}
\includegraphics[width=0.47\textwidth]{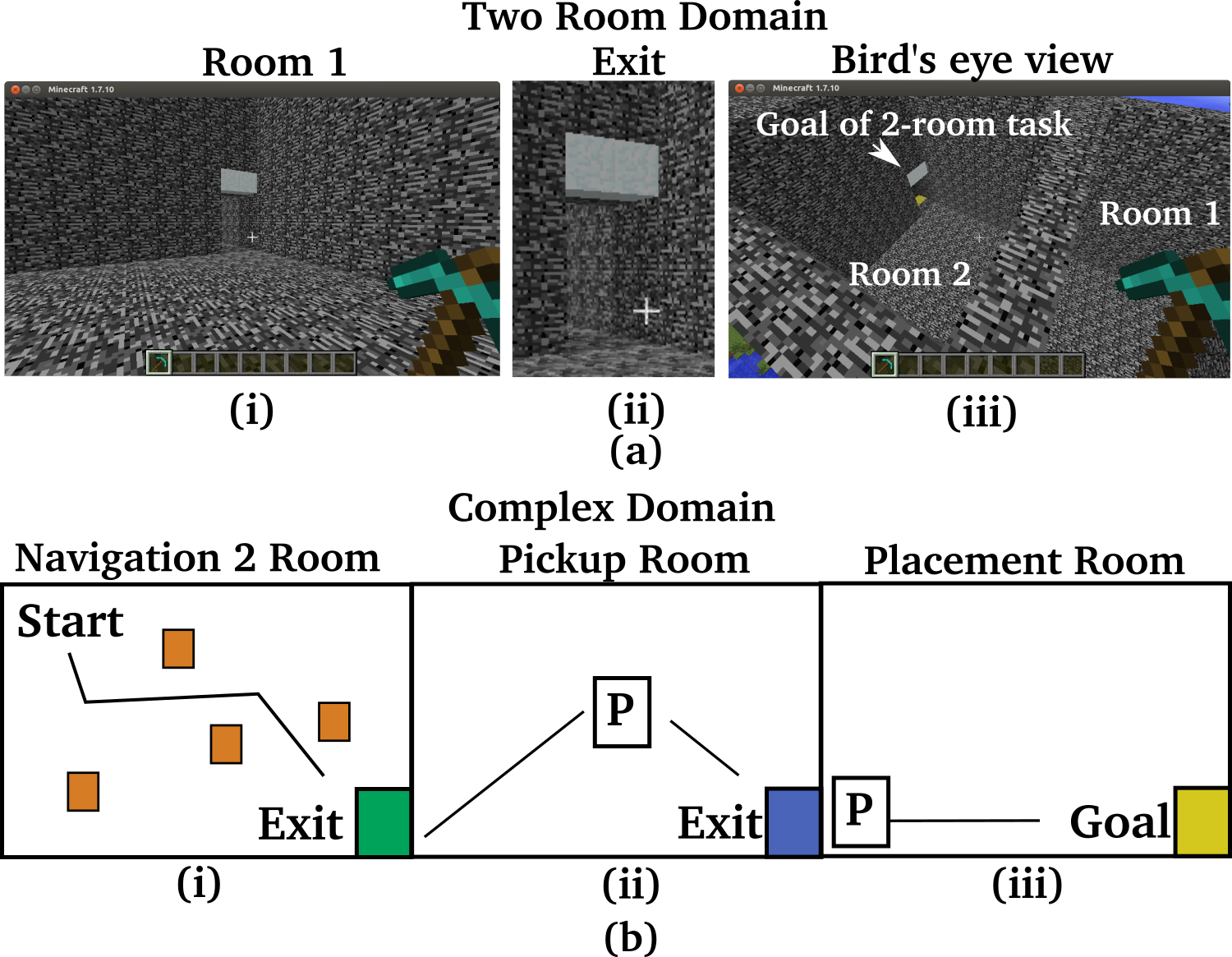} 
\caption{\textbf{Composite domains:} ($a$) The two-room domain and ($b$) the complex domain with three different tasks, ($i$) navigation, ($ii$) pickup and ($iii$) placement}
\label{fig:domains_complex}
\end{center}
\end{figure}

\textbf{Two room Domain: } This domain consists of two-rooms (Figure \ref{fig:domains_complex}$a(iii)$). The first room is shown in Figure \ref{fig:domains_complex}$a(i)$ with its corresponding exit (Figure \ref{fig:domains_complex}$a(ii)$). Note that the exit of the first room is not identical to the exit of the navigation $1$ domain (Figure \ref{fig:domains_dsn}$a$). The second room contains a goal (Figure \ref{fig:domains_complex}$a (iii)$) that is the same as the goal of the navigation $1$ domain (Figure \ref{fig:domains_dsn}$a$). The agent's available action set consists of the primitive movement actions and the Navigate $1$ DSN. \\

\textbf{Skill Reusability/Knowledge Transfer: } We trained the H-DRLN architecture as well as the vanilla DQN on the two-room domain. We noticed two important observations. 
\textbf{(1)} The H-DRLN architecture solves the task after a single epoch  and generates significantly higher reward compared to the vanilla DQN. This is because the H-DRLN makes use of knowledge transfer by \textit{reusing} the DSN trained on the one-room domain to solve the two-room domain. This DSN is able to identify the exit of the first room (which is different from the exit on which the DSN was trained) and navigates the agent to this exit. The DSN is also able to navigate the agent to the exit of the second room and completes the task. The DSN is a temporally extended action as it lasts for multiple time steps and therefore increases the exploration of the RL agent enabling it to learn to solve the task faster than the vanilla DQN. \textbf{(2)} After $39$ epochs, the vanilla DQN completes the task with  $50\%$ success percentage. This sub-optimal performance is due to wall ambiguities, causing the agent to get stuck in sub-optimal local minima. After the same number of epochs, the agent completes the task using the H-DRLN with  $76\%$ success.\\

\begin{figure}[b]
\begin{center}
\includegraphics[width=0.35\textwidth]{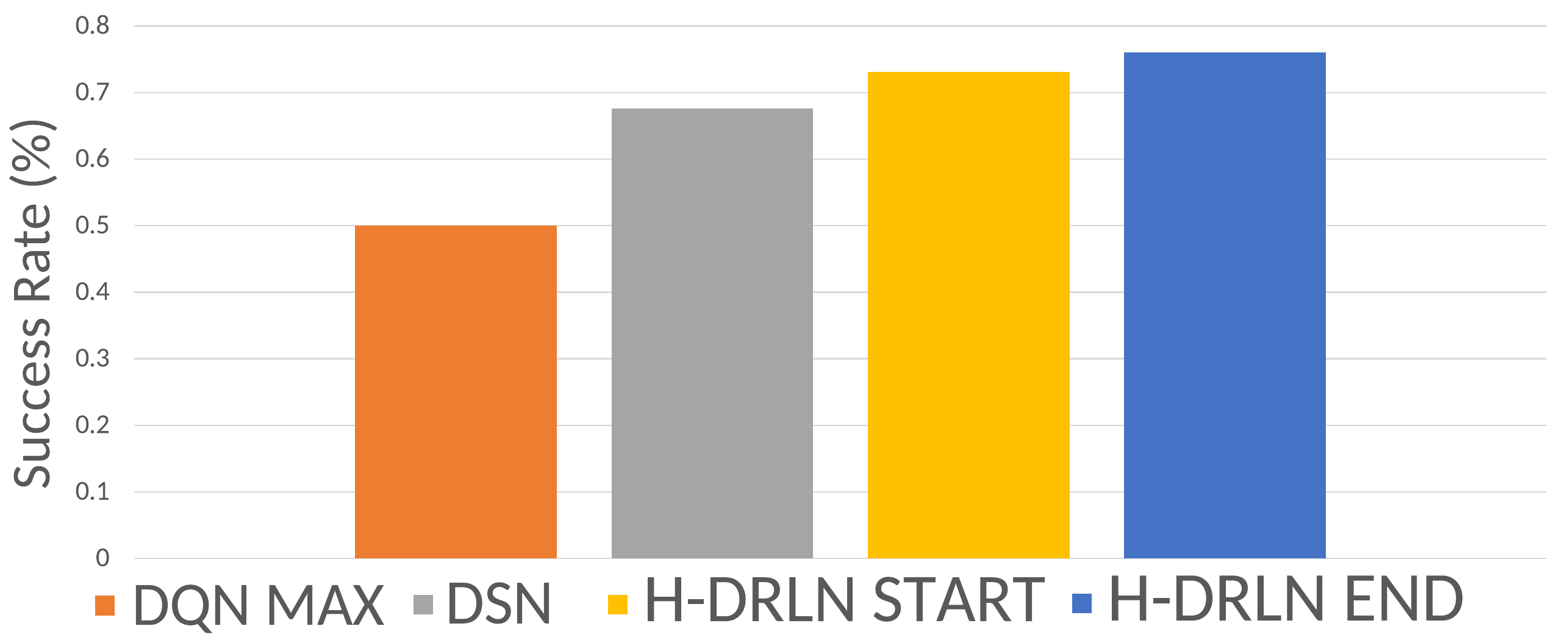} 
\caption{Two room domain \textbf{success percentages} for the vanilla DQN, the single DSN, the H-DRLN after a single epoch (START) and in the last epoch (END).}
\label{fig:bars}
\end{center}
\end{figure} 

\textbf{Knowledge Transfer without Learning:} We then decided to evaluate the DSN (which we trained on the navigation $1$ domain) in the two-room domain \textbf{without} performing any additional learning on this network. We found it surprising that the DSN, without any training on the two-room domain, generated a higher reward compared to the vanilla DQN which was specifically trained on the two-room domain for $39$ epochs. Figure~\ref{fig:bars} summarizes the success percentage comparison between the different architectures in the two-room domain. The vanilla DQN, DSN, H-DRLN START and H-DRLN END had average success percentages of $50\%, 67.65\%, 73.08\%$ and $76\%$ respectively. The DSN performance is sub-optimal compared to the H-DRLN architecture but still manages to solve the two-room domain.  This is an exciting result as it shows the potential for DSNs to identify and solve related tasks without performing any additional learning.\\

\subsection{Training an H-DRLN with a Deep Skill Module}
In this section, we discuss our results for training and utilizing the H-DRLN with a Deep Skill Module to solve the complex Minecraft domain. In each of the experiments in this section, we utilized DDQN to train the H-DRLN and the DDQN baseline unless otherwise stated.\\

\textbf{Complex Minecraft Domain:}
This domain (Figure \ref{fig:domains_complex}$b$) consists of three rooms. Within each room, the agent is required to perform a specific task. Room $1$ (Figure \ref{fig:domains_complex}$b(i)$) is a navigation task, where the agent needs to navigate around the obstacles to reach the exit. Room $2$ (Figure \ref{fig:domains_complex}$b(ii)$) contains two tasks. (1) A pickup task whereby the agent is required to navigate to and collect a block in the center of the room; (2) A break task, where the agent needs to navigate to the exit and break a door. Finally, Room $3$ (Figure \ref{fig:domains_complex}$b(iii)$) is a placement task whereby the agent needs to place the block that it collected in the goal location. The agent receives a non-negative reward if it successfully navigates through room $1$, collects the block and breaks the door in room $2$ and places the block in the goal location in room $3$ (Arrow path in Figure \ref{fig:domains_complex}$b$). Otherwise, the agent receives a small negative reward at each timestep. Note that the agent needs to complete three separate tasks before receiving a sparse, non-negative reward. The agent's available action set are the original primitive actions as well as the DSN's: (1) Navigate $2$, (2) Pickup, (3) Break and (4) Placement.\\

\begin{figure}[h!]
\begin{center}
\includegraphics[width=0.40\textwidth]{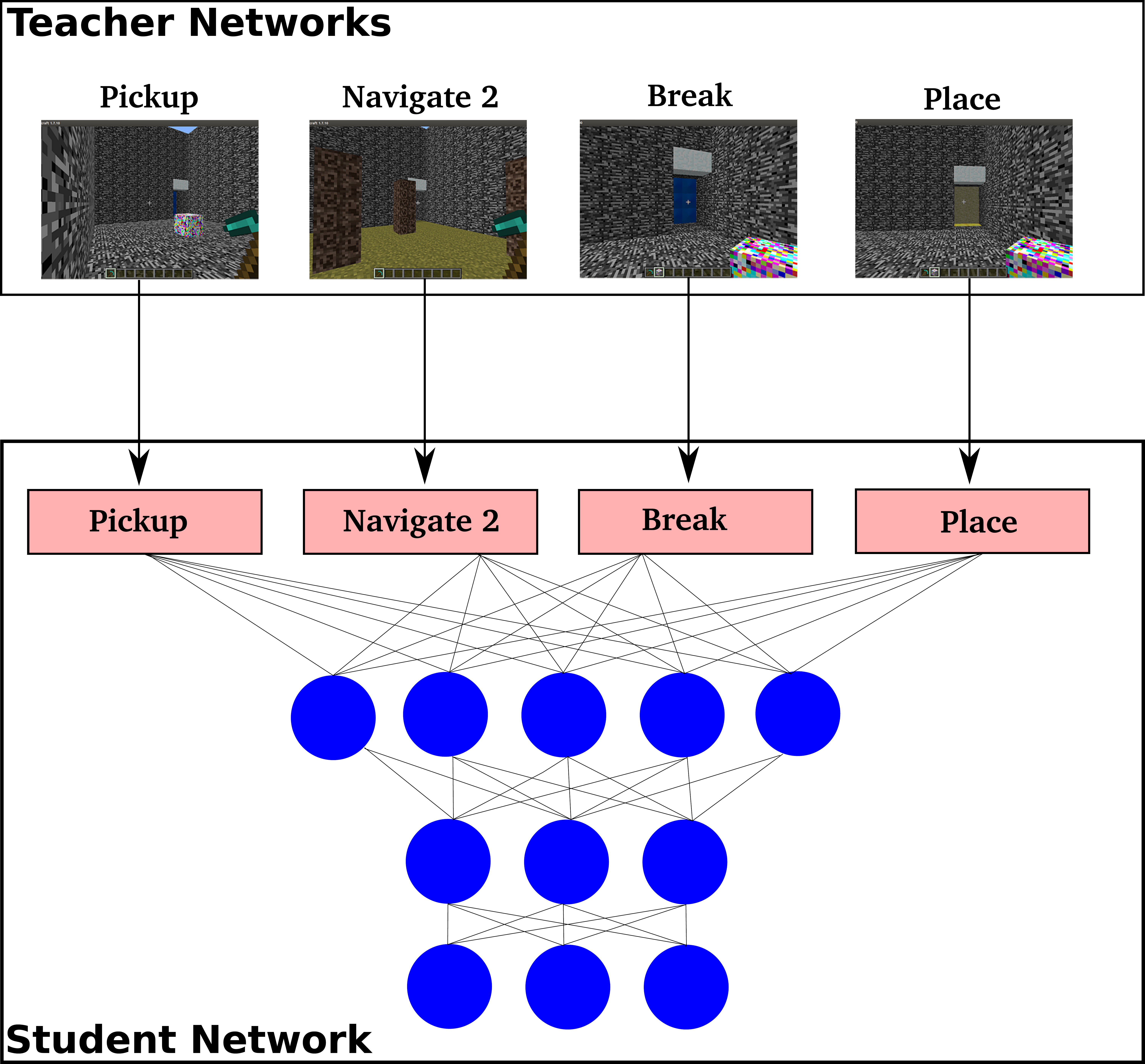} 
\caption{Multi-skill distillation.}
\label{fig:distillation_fig}
\end{center}
\end{figure}

\textbf{Training and Distilling Multiple DSNs:}
As mentioned in the H-DRLN Section, there are two ways to incorporate skills into the Deep Skill Module: (1) DSN Array and (2) Multi-Skill Distillation. For both the DSN array and multi-skill distillation, we utilize four pre-trained DSNs (Navigate $2$, Pickup, Break and Placement). 
These DSNs collectively form the DSN array. For the multi-skill distillation, we utilized the pre-trained DSNs as teachers and distil these skills directly into a single network (the student) using the distillation setup shown in Figure \ref{fig:distillation_fig}, and as described in the Background Section. Once trained, we tested the distilled network separately in each of the three individual rooms (Figure \ref{fig:domains_dsn}$b-d$). The performance for each room is shown in Table \ref{table:distilled} for temperatures $\tau=0.1$ and $\tau=1$. The high success percentages indicate that the agent is able to successfully complete each task using a single distilled network. In contrast to policy distillation, our novelty lies in the ability to, not only distil skills into a single network, but also learn a control rule (using the H-DRLN) that switches between the skills to solve a given task.

\begin{table}[h]
\centering
\begin{tabular}{| c | c | c | c |}
\hline
Domain &  $\tau = 0.1$ & $\tau = 1$ & Original DSN \\ \hline
Navigation & 81.5 & 78.0 & 94.6  \\ \hline
Pick Up & 99.6  & 83.3 & 100\\ \hline
Break & 78.5  & 73.0 & 100\\ \hline
Placement & 78.5  & 73.0 & 100\\ \hline
\end{tabular}
\caption{The \textbf{success $\%$} performance of the distilled multi-skill network on each of the four tasks (Figures \ref{fig:domains_dsn}$b-d$). }
\label{table:distilled}
\end{table}

\begin{figure}
\begin{center}
\includegraphics[width=0.47\textwidth]{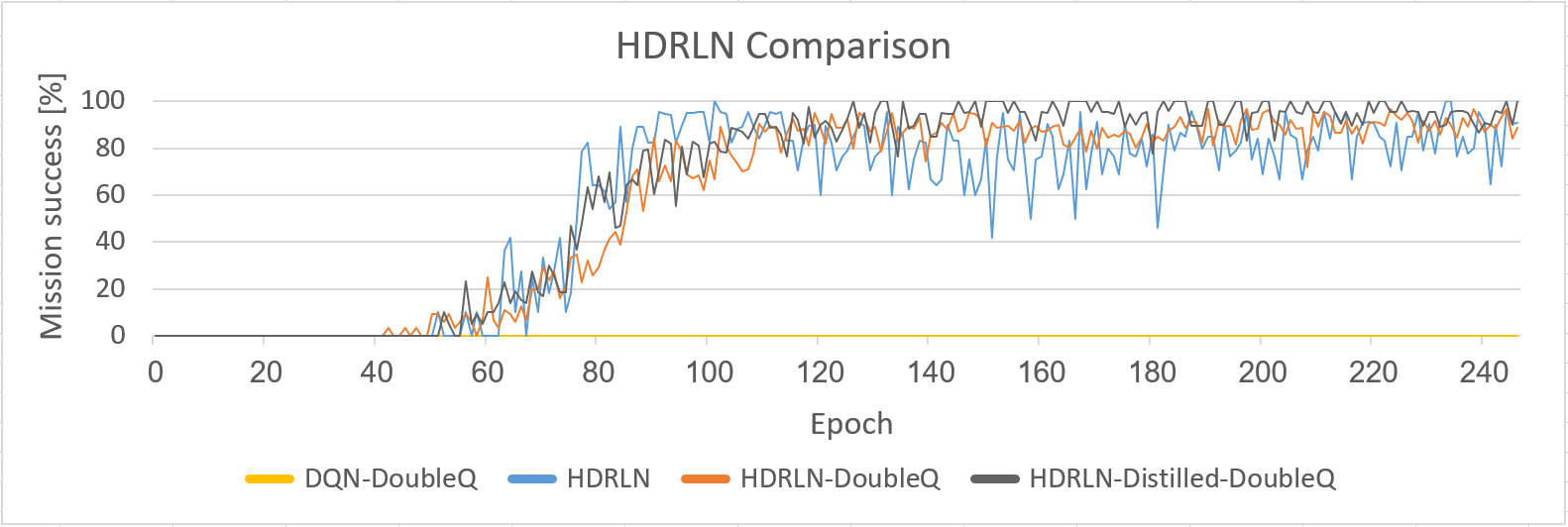} 
\caption{The success $\%$ \textbf{learning curves} for the (1) H-DRLN with a DSN array (blue), (2) H-DRLN with DDQN and a DSN array (orange), and (3) H-DRLN with DDQN and multi-skill distillation (black). This is compared with (4) the DDQN baseline (yellow).}
\label{fig:learningcurves}
\end{center}
\end{figure}

\textbf{Training the H-DRLN:}
We now show results for training the (1) H-DRLN with a DSN array, (2) H-DRLN with DDQN and a DSN array and (3) H-DRLN with DDQN and a distilled multi-skill network (with $\tau=0.1$). This is compared to (4) a DDQN baseline. The learning curves can be seen in Figure \ref{fig:learningcurves}. We performed these trials 5 times for each architecture and measured success rates of $85\pm10\%$, $91\pm4\%$ and $94\pm4\%$ ($mean\% \pm std$) for the H-DRLN, H-DRLN with DDQN and H-DRLN with DDQN and a distilled multi-skill network respectively. To calculate these values we averaged the success percentages for the final $10$ epochs. Note that the distilled H-DRLN has a higher average success rate and both H-DRLN's with DDQN have lower variance. The DDQN was unable to solve the task. This is due to a combination of wall ambiguities (as in the two room domain) and requiring more time to learn. The H-DRLN is able to overcome ambiguities and also learns to reuse skills. We also trained the DDQN with intrinsic rewards which enabled it to solve the task. However, this required a significantly larger amount of training time compared to the H-DRLN and the result was therefore omitted.\\

\begin{figure}[h!]
\begin{center}
\includegraphics[width=0.47\textwidth]{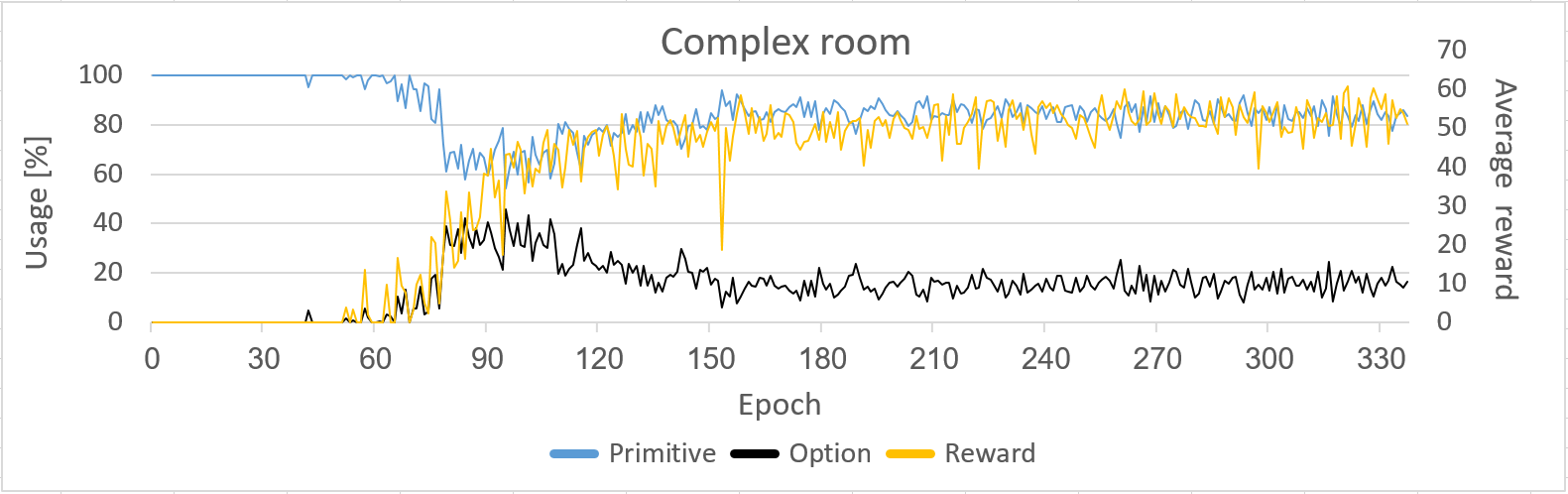} 
\caption{\textbf{Skill usage} $\%$ in the complex domain during training (black). The primitive actions usage $\%$ (blue) and the total reward (yellow) are displayed for reference.}
\label{fig:optionselection}
\end{center}
\end{figure}

\textbf{Skill usage:}
Figure \ref{fig:optionselection} presents the usage $\%$ of skills by the H-DRLN agent during training. We can see that around training epoch $50$, the agent starts to use skills more frequently (black curve). As a result, the H-DRLN agent's performance is significantly improved, as can be seen by the increase in reward (yellow curve). After epoch $93$, the agent's skill usage reduces with time as it needs to utilize more primitive actions. This observation makes sense, since planning only with skills will yield a sub-optimal policy if the skills themselves are sub-optimal. However, planning with both primitive actions and skills always guarantees convergence to an optimal policy (utilizing only primitive actions in the worst-case) \cite{Mann2013}. In our case, the skills that were trained on the one-room domains helped the agent to learn in the complex domain but were sub-optimal  due to small changes between the one-room domains and the complex domain. Thus, the agent learned to refine his policy by using primitive actions. To conclude, Figures \ref{fig:learningcurves} and \ref{fig:optionselection} tell us that, while skills are used approximately $20\%$ of the time by the final H-DRLN policy, they have a significant impact on accelerating the agent's learning capabilities.

\section{Discussion}
We presented our novel Hierarchical Deep RL Network (H-DRLN) architecture. This architecture contains all of the basic building blocks for a truly general lifelong learning framework: (1) Efficient knowledge retention via multi-skill distillation; (2) Selective transfer using temporal abstractions (skills); (3) Ensuring interaction between (1) and (2) with the H-DRLN controller. We see this work as a building block towards truly general lifelong learning using hierarchical RL and Deep Networks.\\

We have also provided the first results for learning Deep Skill Networks (DSNs) in Minecraft, a lifelong learning domain. The DSNs are learned using a Minecraft-specific variation of the DQN \cite{Mnih2015} algorithm. Our Minecraft agent also learns how to reuse these DSNs on new tasks by the H-DRLN. We incorporate multiple skills into the H-DRLN using (1) the DSN array and (2) the scalable distilled multi-skill network, our novel variation of policy distillation.\\

In addition, we show that the H-DRLN provides superior learning performance and faster convergence compared to the DDQN, by making use of skills. Our work can also be interpreted as a form of curriculum learning \cite{bengio2009curriculum} for RL. Here, we first train the network to solve relatively simple sub-tasks and then use the knowledge it obtained to solve the composite overall task. We also show the potential to perform knowledge transfer between related tasks without any additional learning. This architecture also has the potential to be utilized in other 3D domains such as Doom \cite{Kempka2016} and Labyrinth \cite{Mnih2016asynchronous}.\\

Recently, it has been shown that Deep Networks tend to implicitly capture the hierarchical composition of a given task \cite{Zahavy2016}. In future work, we plan to utilize this implicit hierarchical composition to learn DSNs. In addition, we aim to (1) learn the skills online whilst the agent is learning to solve the task. This could be achieved by training the teacher networks (DSNs), whilst simultaneously guiding learning in the student network (our H-DRLN); (2) Perform online refinement of the previously learned skills; (3) Train the agent in real-world Minecraft domains.

\section*{Acknowledgement}
This research was supported in part by the European Community’s Seventh Framework Programme (FP7/2007-2013) under grant agreement 306638 (SUPREL) and the Intel Collaborative Research Institute for Computational Intelligence (ICRI-CI).

\bibliographystyle{aaai}
\bibliography{tmann}

\end{document}